\pdfoutput=1

\documentclass[11pt]{article}

\usepackage[]{EMNLP2023}

\usepackage{listings}
\usepackage{graphicx}

\usepackage{times}
\usepackage{latexsym}

\usepackage[T1]{fontenc}

\usepackage[utf8]{inputenc}
\usepackage{svg}

\usepackage{microtype}

\usepackage{caption}
\usepackage{subcaption}

\usepackage{inconsolata}

\newcommand{\gt}{$^{*}$}
\newcommand{\stanf}{$^{\dagger}$}

\title{A Material Lens on Coloniality in NLP}

\author{William Held \gt\stanf \hspace{1.5em}
        Camille Harris \gt \hspace{1.5em}
        \textbf{Michael Best}\gt \hspace{1.5em}
        \textbf{Diyi Yang}\stanf \\
        \gt Georgia Institute of Technology, \stanf Stanford University \\
        \texttt{\small wheld3@gatech.edu}, \texttt{\small diyiy@stanford.edu}
}

\begin{document}
\maketitle
\begin{abstract}
Coloniality, the continuation of colonial harms beyond "official" colonization, has pervasive effects across society and scientific fields. Natural Language Processing (NLP) is no exception to this broad phenomenon. In this work, we argue that coloniality is implicitly embedded in and amplified by NLP data, algorithms, and software. We formalize this analysis using Actor-Network Theory (ANT): an approach to understanding social phenomena through the network of relationships between human stakeholders and technology. We use our Actor-Network to guide a quantitative survey of the geography of different phases of NLP research, providing evidence that inequality along colonial boundaries increases as NLP builds on itself. Based on this, we argue that combating coloniality in NLP requires not only changing current values but also active work to remove the accumulation of colonial ideals in our foundational data and algorithms.
\end{abstract}

\section{Introduction}
\begin{quote}
    {
    ``\textit{Coloniality\dots refers to long-standing patterns of power that emerged as a result of colonialism, but that define culture, labor, intersubjective relations, and knowledge production well beyond the strict limits of colonial administrations}'' \\\phantom{ab}--- \textbf{Nelson Maldonado-Torres} (\citeyear{maldonado2007coloniality})
    }
\end{quote}
While European colonization, the sovereignty of European nations over non-European nations, has been dismantled on paper in most of the world\footnote{There are still 17 colonies governed by the United States, the United Kingdom, and France.}, it still mars modern society. \textbf{Coloniality}, the continuation of harms against the colonized and centering of the ideals of colonizing powers, is pervasive across geographic regions, intellectual domains, and sectors of society~\citep{said1978orientalism, smith1999decolonizing, quijano2000coloniality, boshoff2009neo, chakrabarti2010materials, wengraf2018extracting, alvarado2021decolonial, ferdinand2021decolonial}, including AI broadly~\citep{mohamed2020decolonial, birhane2020algorithmic}.

\begin{figure*}[t]
\centering
   \begin{subfigure}[b]{0.415\textwidth}
   \includegraphics[width=1\textwidth,trim={0cm 5.5cm 13.1cm 3.8cm},clip]{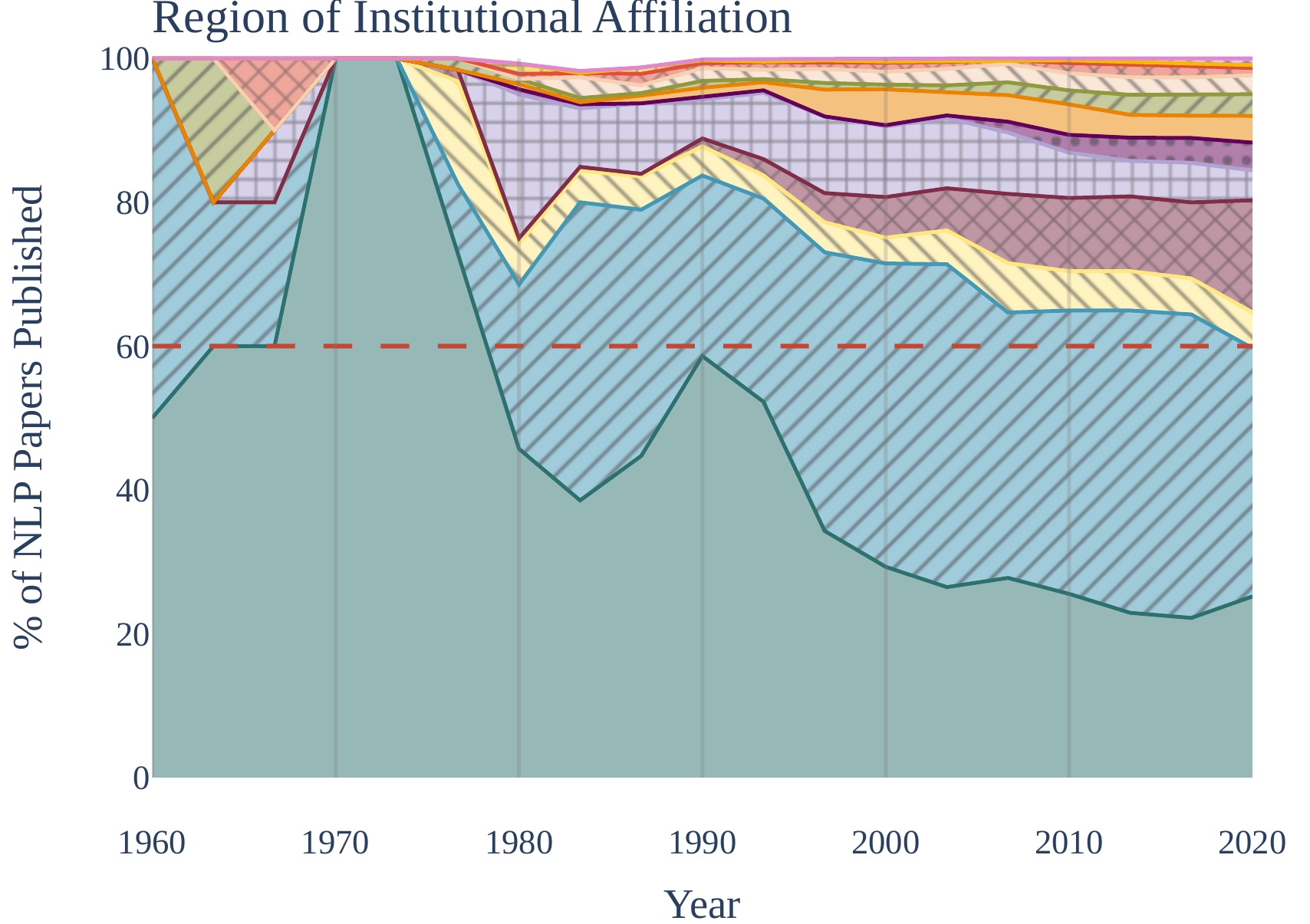}
\end{subfigure}
\hfill
\begin{subfigure}[b]{0.575\textwidth}
   \includegraphics[width=1\textwidth,trim={2cm 3.3cm 0cm 5cm},clip]{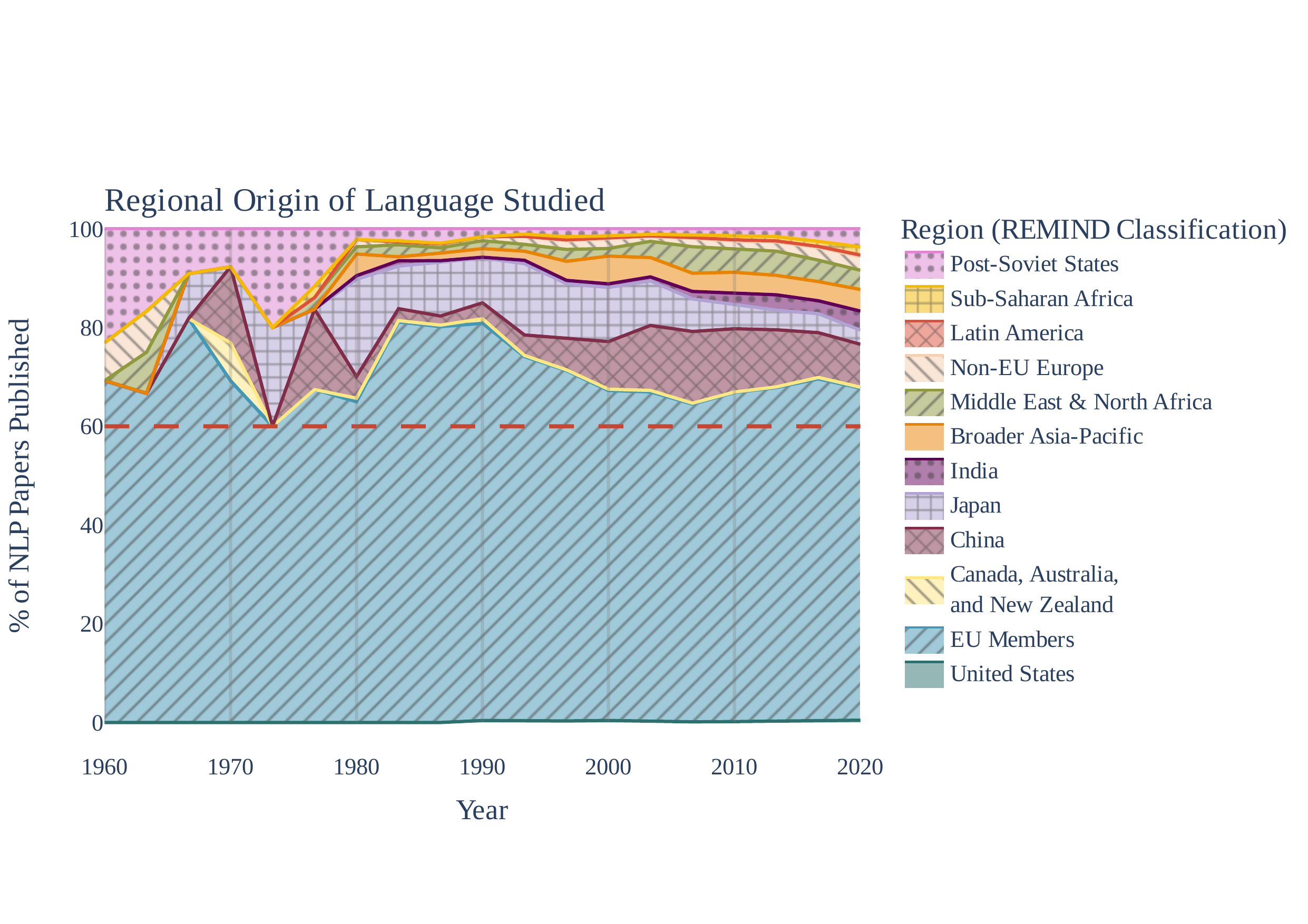}
\end{subfigure}
\caption{Natural Language Processing (NLP) research grouped by the regional affiliation of authors and regional origin of languages studied using the REMIND region classification~\citep{remind}. Across 7 decades of NLP research, institutions based in Western Europe and British settler-colonial states have always published over 60\% of NLP papers. While science broadly follows this dynamic~\citep{index2014introducing}, in NLP this shapes the research itself to be Eurocentric, across the same period over 60\% of research focused on Western European languages.}
\label{fig:headline}
\end{figure*}

Language studies specifically have a stark colonial history. Up until the 20th century, the linguistics institutions of Non-European languages were led by European scholars, often supported directly by colonial regimes~\citep{errington2001colonial} which suppressed the use of those same languages by their native speakers~\citep{wa1992decolonising}.  Prior works have argued that NLP continues to perpetuate methodological coloniality, especially in research on African and Indigenous Languages~\cite{bird2020decolonising, schwartz-2022-primum, ogunremi2023decolonizing, mager2023ethical}.

These works also show coloniality is not simply a matter of linguistic diversity. As illustrated in Figure \ref{fig:headline}, both languages studied \emph{and} researcher affiliations reflect Western Eurocentrism. The study of new languages can often reflect the shifting interests of existing powers, called interest convergence~\citep{bell1980brown, ogbonnaya2020critical}. For example, in times of geopolitical conflict with the US, languages of what was then the Soviet Union (1960-1980)~\citep{leon2021automating} and later the Middle East (2000-2010)~\citep{farghaly2009arabic} experienced a surge in research, without a proportional increase in the representation of researchers affiliated with those regions.

As many believe that NLP technologies are on the precipice of enacting major societal~\citep{bommasani2021opportunities} and economic change~\citep{eloundou2023gpts}, understanding inequity in people's agency to shape NLP is of increasingly broad importance. Otherwise, systems such as large language models (LLMs) are likely to expand harms such as algorithmic discrimination, exclusion, and stereotyping~\citep{bender2021dangers, weidinger2022taxonomy, hovy2016social, cheng-etal-2023-marked}.

In this paper, we focus on how coloniality can be connected to unstructured data, annotations, models and software in NLP. In turn, these technologies reinforce and amplify coloniality beyond the social systems that created them. This view of technology and objects as agents that resist or create change in sociotechnical systems is called "\textbf{materiality}"~\citep{bridges, appadurai1988social, keane2003semiotics, hodder2012entangled}.

\begin{figure*}[ht]
    \centering
    \includegraphics[width=0.98\textwidth,trim={0cm 0.1cm 0cm 0.1cm},clip]{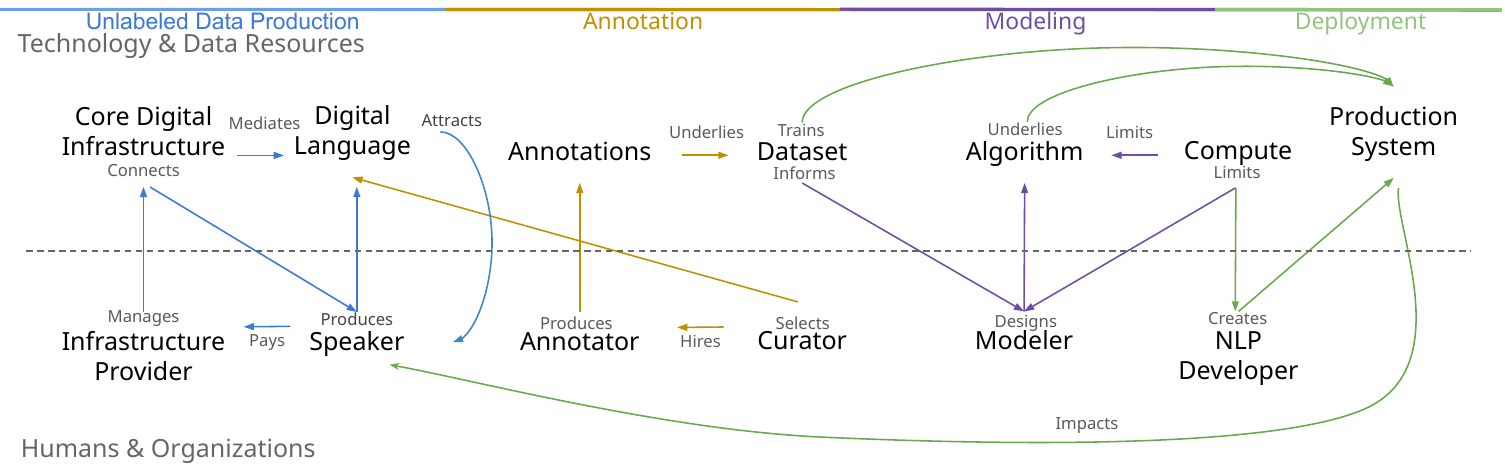}
        \caption{The Actor-Network of NLP Development and Deployment. Unlabeled data production process documented in \citet{pimienta2009twelve} and \citet{kornai2013digital}. Annotation process documented in \citet{bender-friedman-2018-data}. Modeling process documented in \citet{ethayarajh2020utility} and \citet{hooker2021hardware}. Deployment and impacts documented in \citet{hovy2016social} and  \citet{mitchell2019model}.}
    \label{ant_graph}
\end{figure*}

To formally guide our analysis, we leverage Actor-Network theory~\citep{latour2005reassembling} (ANT), a conceptual framework that allows us to relate people and technology jointly. ANT is widely used in Science and Technology Studies, including for digital and data-driven technologies~\citep{bowker1999sorting, stanforth2006using, kumar2013mobile, frauenberger2019entanglement, selbst2019fairness}. In ANT, both human and technological actors are connected via their relationships in a so-called Actor-Network. This can then be used to identify people, organizations, and technologies that stabilize and destabilize power dynamics in the network.

To support understanding of coloniality in NLP, we contribute the following:

\begin{enumerate}
    \setlength{\itemsep}{0.025in}
    \setlength{\parskip}{0pt}
        \setlength{\parsep}{0pt}
    \item \textbf{Survey of the NLP Actor-Network:} We survey prior literature that documents technological artifacts and their production process. We then synthesize these works into an Actor-Network for NLP research and deployment.

    \item \textbf{Interpreting the Material Expression of Coloniality in NLP:} We study the direct effects of coloniality on technological actors with this Actor-Network. We then analyze how technologies stabilize the social power structures of the field even as social pressures and technical capabilities evolve.
    
    \item \textbf{Quantitative and Qualitative Studies of Coloniality in NLP:} Finally, we use our Actor-Network to guide a quantitative analysis of a corpus of all *CL publications up to September 2022, described in Section \ref{app:corpus}, to measure inequality as NLP at different stages of NLP development. We use qualitative case studies to highlight that our quantitative analysis is an underestimate of impacts since geographic borders and languages are only surface-level measures.
\end{enumerate}

\citet{liboiron2021pollution} notes that coloniality "\emph{can be maintained by good intentions and even good deeds.}" Through our focus, we argue that these material aspects of coloniality have pervasive downstream effects on NLP researchers, even if we assume good intentions. Therefore, we must actively work to reshape these technological foundations of NLP according to communal values.

\paragraph{Author Positionality} 
This work began as a reflection on the first author's privilege as a White Standard American English-speaking NLP researcher from the US working on dialects and languages of which they are not a native speaker. The power structures documented benefit not just an anonymous other but directly benefit the authors of this work who all operate from US-based institutions. This work should be viewed as support of the foundational theories in NLP it is built upon, especially \citet{mohamed2020decolonial}, \citet{bird2020decolonising}, \citet{ogunremi2023decolonizing}, and \citet{mager2023ethical}.

\section{Building The Actor-Network}
\label{sec:build}
To develop our Actor-Network separately from our analysis of coloniality, we survey works that document the NLP data and algorithm development process; placing them together in an Actor-Network in Figure \ref{ant_graph}. First, we describe the factors which lead speakers to produce unlabeled raw language data. Second, we document how raw data is curated and annotated as a resource for NLP systems. Third, we draw connections between annotated data and the capabilities and development process of NLP algorithms. Finally, we look at how the impacts of NLP research are realized through product deployment.

\subsection{Unlabeled Digital Data Production}
The foundation of NLP is data upon which datasets are constructed, systems are trained and tested, and products operate. Often, this is sourced from the Internet and it must always be somehow digital.

During early Internet adoption, UNESCO commissioned several studies to understand digital language use, producing a thorough registry of the requirements for a language to be used online. For content to be accessible in most NLP datasets, Internet Service Provider infrastructure must be available. This creates a major barrier, which is largely dependent on geography~\citep{pimienta2009twelve}. Even for offline textual data, the language must be supported by both hardware, such as keyboards~\citep{mullaney2017chinese}, and software, such as encoding schemes~\citep{dombrowski2020preparing}, to allow for the production and consumption of digital text~\citep{paolillo2005measuring} or be scanned and extracted from non-digital resources~\citep{rijhwani-etal-2020-ocr}.

Beyond access, social factors influence discrepancies in the production of language data~\citep{hovy-yang-2021-importance,hovy2016social}. The strong network effects of content production make the size of existing online communities a stronger predictor of language vitality online than raw population numbers~\citep{kornai2013digital}. This creates a cycle of content production once a critical mass of information is available in a language. As speakers are attracted to the web by existing websites, they themselves are likely to become content producers. Translation does not resolve this issue, as the quality of software~\citep{kornai2013digital} and the investment in digital literacy~\citep{hargittai2001second} create their own barriers to adoption and content production.

\subsection{Annotated Datasets as Stores of Value}
In addition to unlabeled data, NLP systems often rely on curated examples of language with additional annotations. From the outset, the process of curation is affected by the raw data distribution. As the process of data curation and annotation is repeated, curators often use the same distribution of data~\citep{raji2021ai} leaving others systematically excluded~\citep{hovy2016social}. 

In developing standards of documentation for NLP datasets, \citet{bender-friedman-2018-data} carefully connect the social aspects of language dataset creation to the dataset itself. This is an inherently materialist perspective --- this documentation is necessary because the resource propagates these social forces. Speakers who produce raw data, curators who select raw data, and annotators who produce individual annotations, all impact the resulting datasets. For any language, only a subset of the speaker population participates in this process; many who may be impacted by resulting systems are excluded. As such, data statements treat the identities, assumptions, and goals of those involved as a fundamental property of the digital artifact. 

Curators are especially notable in this process as, beyond curation, they often recruit annotators and design annotation guidelines. This dynamic has increased as crowdwork has become prevalent in the field~\citep{shmueli-etal-2021-beyond}, especially for large-scale Reinforcement Learning From Human-Feedback~\citep{ouyang2022training, touvron2023llama}. This leaves curators as the primary authority on the guidelines, labels, and formats that underly annotated datasets. By deciding how the near infinite space of language is compressed into fixed categories, the ideals implicit in this structure are applied not only to the data that they have collected but to all data they feel it represents~\citep{todd2016indigenous}.

\subsection{Social Incentives in Modeling Work}
One major difference between the closely related fields of NLP and Computational Linguistics is the focus on empirical results. A famous, albeit of contested exact phrasing, quote from an NLP pioneer succinctly captures this value difference:  "\emph{Every time I fire a linguist, the performance of the system goes up.}"~\citep{hirschberg1998every}. This shift in priorities was shaped by required participation in shared tasks during the MUC~\citep{chinchor-1991-muc, sundheim-1995-overview} and ATIS~\citep{hemphill-etal-1990-atis} workshops in the late 1980's and early 1990's. These were central events in the field at the time, since grantees funded by the US Defense Advanced Research Projects Agency (DARPA) were encouraged to attend~\citep{anderson-etal-2012-towards}. As participants in these events established and led future areas of NLP, benchmarks and shared tasks became a major driver of NLP model development~\citep{strassel2006corpus, paroubek2007principles, wang2018glue, raji2021ai, srivastava2023beyond}. 

However, defining and ranking by "performance" requires reducing the qualitative features of a model to a minimal set of sortable metrics. Inevitably, some aspects of true performance are lost. \citet{ethayarajh2020utility} use game theory to argue that this rationally dictates the priorities of model developers. Optimizing widely reported dataset metrics can lead to publicity and career rewards. These incentives can override other dimensions of practical utility to speakers~\citep{ruder-etal-2022-square}. Practically, modeling researchers report data availability as a major factor in deciding research directions~\citep{thakkar2022ml} due to the relatively low prestige and high difficulty of data work~\citep{sambasivan2021everyone}.

Furthermore, \citet{hooker2021hardware} highlights that algorithm design has a second non-theoretical limiting factor --- hardware.  Algorithms that are mathematically or linguistically sound often fail if they are not well-supported by existing hardware. Even the now dominant neural networks exploded in popularity primarily after high quality frameworks for GPU support were released~\citep{bergstra2010theano} and \emph{Large} Language Models developed alongside hardware to support them~\citep{saphra2023tragedy}.

Benchmarks and hardware, both non-human actors, thereby influence the design process of model developers. These constraints increase as NLP becomes more machine learning driven~\citep{manning1999foundations, sevilla2022compute} and especially reliant on larger datasets~\citep{gao2020pile, birhane2022values,wang2018glue} and models~\citep{strubell2019energy, brown2020language, peterEnvironment, patterson2021carbon}.

\subsection{NLP Deployment Realizes Impact}
Thus far we have focused on how social forces shape NLP, but NLP research leads to products and tools that impact non-NLP researchers and broader society. 
Similar to data statements, \citet{mitchell2019model} provides guidelines for documenting how Machine Learning model development should dictate deployment. Released models, their functionality and their intended use cases all influence what products are possible.

When multiple models are integrated inside of products, they transform these technical functionalities into non-technical impacts on those who use the product or to whom it is applied. As noted in \citet{hovy2016social}, these systems impact the broader community that uses a language, regardless of whether they were included in the annotation or algorithm development process. 

These applications are not asocial. They can have positive impacts such as creating economic opportunities~\citep{brynjolfsson2017can, brynjolfsson2019does}, informing public policy~\citep{jurgens2017incorporating}, and aiding education~\citep{loukina2019many, wang-demszky-2023-chatgpt} or negative impacts through dual use such as harming privacy~\citep{jurgens2017writer}, distributing harmful knowledge~\citep{shaikh-etal-2023-second}, and generating misinformation~\citep{zhou2023synthetic}. Through deployment, inequities in the social systems that shape NLP are replicated through technological harms and allocative inequities.

\section{Quantitative Actor-Network Analysis}
\label{app:corpus}
While theoretical analysis of the Actor-Network itself can create interesting hypotheses, one promising feature of using materiality as a lens is that it can be meaningfully measured. At the scale of the entire field, we believe there are two key traits which exist in all NLP works: the social environment which shapes the research and the language the research studies. To avoid generalizing about authors by their identity, we approximate "social environment" via the institutional affiliations and, by proxy, regional affiliation of each work. Admittedly, both regional affiliation and broadly defined "languages" are coarse and likely to miss nuances, we explore specific material effects of coloniality through qualitative case studies in Section \ref{sec:case}.

For all quantitative results in this work, we rely on analysis on the full set of *CL papers released on the ACL Anthology up until September 2022 from the ACL-OCL-Corpus~\citep{rohatgi2023acl}. While ACL-OCL offers full text and abstracts for each papers, the author affiliations are not indexed. Therefore, we download and reparse all PDFs using the S2ORC-Doc2Json tool~\citep{lo-wang-2020-s2orc}, including an additional 4,332 papers whose PDFs were hosted on domains other than aclanthology.org where full-text was not available in the original corpus. The result is a set of 58k papers for which we have full-text, author affiliation mappings, and paper language mappings.

\subsection{Author Affiliation Methodology}
\label{app:aff}
In an ideal world, each paper would be mapped to a location by the authors themselves. This was often the practice prior to the prevalence of email when mailing addresses were more relevant to academic communication. However, in practice, authors often report different aspects of information in each paper. We use the following cascade of rules to map each author on a paper to their affiliation:
\begin{enumerate}
    \item If the author lists an email address with a country-level top-level domain, e.g. "cn" for China, we assign the affiliation as that country.
    \item If the author lists a country name explicitly, we assign the affiliation as that country.
    \item If the author lists a postal code, we get all possible cities associated with the postalcode from \href{https://www.geonames.org/}{GeoNames}. If the author also lists exactly one of these cities, we assign the affiliation as the country which contains the matched city and postal code pair.
    \item If the author lists an University email address with a domain in the list from ~\citet{citationgaps}, we assign the affiliation as the country associated with that University.
    \item If the author lists an institution name which has at most one character difference from an institution name in the list from ~\citet{citationgaps}, we assign the affiliation as the country associated with that University.
\end{enumerate}

We evaluate these rules in the order listed above, prioritizing rules which are less likely to be affected by noise from the GrobID~\citep{lopez2009grobid} parsing tool or to incorrectly map affiliations based on text matching alone.

\subsection{Language Mention Methodology}
\label{app:lang_mentions}
Following \citet{joshi-etal-2020-state}, we first search the plain text of all papers in our corpus for mentions of each language listed in their lang2tax file. Upon manual inspection, we find a reasonable number of spurious matches such as homonyms of the language, \textit{"...how they work (Serrano and Smith, 2019..."} being counted for the Serrano language, or incidental mentions of languages, "Unlike English..." being counted for English. 

To handle this, we filter all mentions using GPT-3.5-Turbo as a filtering tool. We sample 5 sentences which include a full-text language match and prompt the LLM to produce its own set of languages which are the main focus.
\begin{lstlisting}
{
      "role": "system",
      "content": "You are a Natural Language Processing expert carefully studying papers from ACL. On each line, only return valid Python set.",
},
{
      "role": "user",
      "content": str(
  'What are the primary languages of interest of a paper with these sentences? Ignore languages that are only mentioned in passing, for example mentions like "Unlike English": should not lead to English being included in the set. \n Sentences: '
  + "\n".join(LANG_MENTIONS)])
      ),
},
\end{lstlisting}

To avoid hallucinations from the language model, we use the intersection between the original set of languages explicitly mentioned in the text and the set of languages returned by GPT-3.5-Turbo. This means that the precision is, at worst, the same as the method from \citet{joshi-etal-2020-state}. Notably, this likely undercounts the concentration of research in English, as many papers may focus on English without following the so-called "Bender Rule" of stating the language of interest regardless~\citep{bender2011achieving}.

\begin{figure*}[ht]
    \centering
    \includegraphics[width=1\textwidth,trim={2cm 2cm 1cm 3cm},clip]{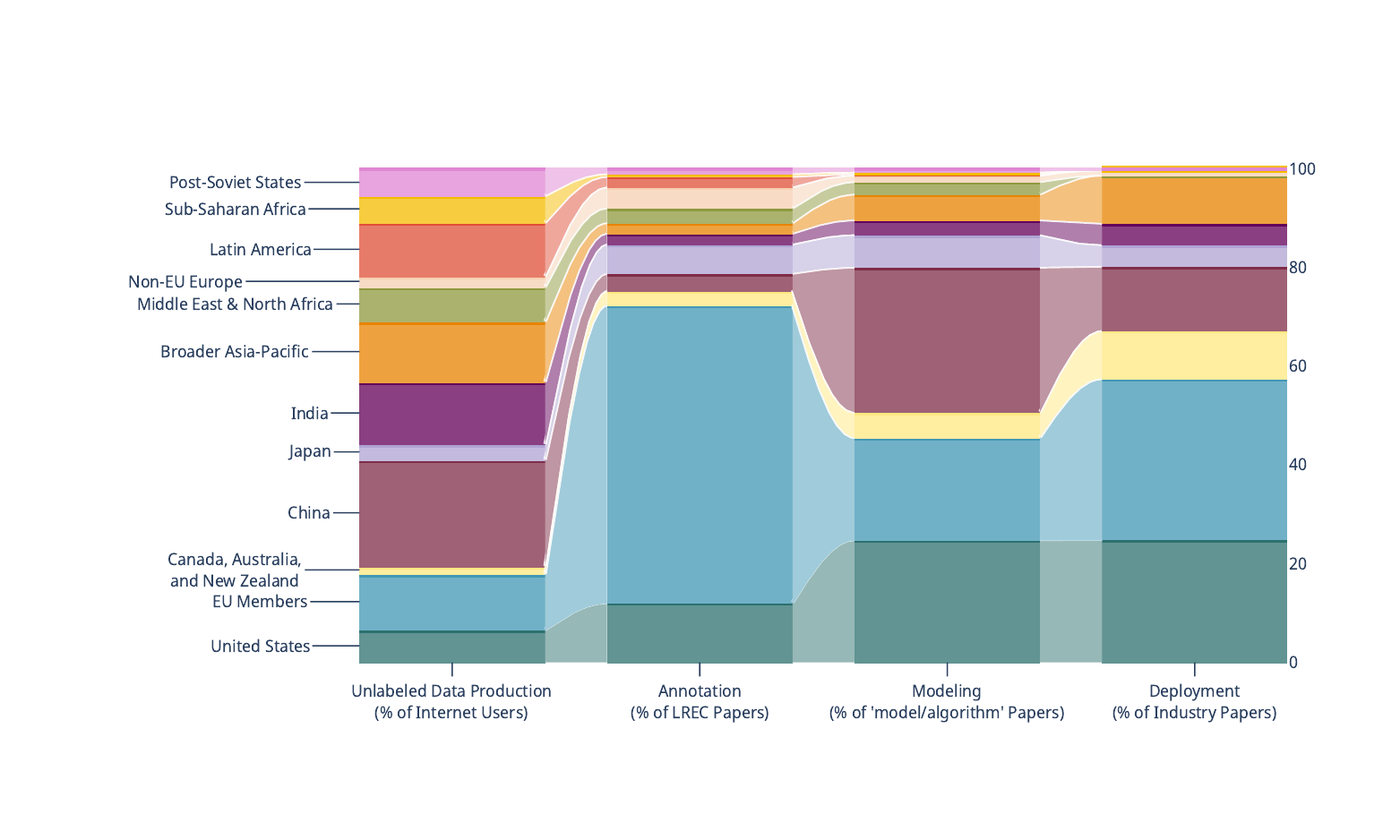}
        \caption{Geographic distribution of the people involved with each phase of the NLP pipeline, from raw data to deployed model, in papers since 2015. Representation of heavily colonized regions, such as Latin America and Sub-Saharan Africa, dwindles throughout the NLP pipeline. See Section \ref{app:aff} and Section \ref{app:class} respectively for details on how we extract authors country affiliations and sample papers for each phase.}
    \label{ant_maps}
\end{figure*}

\subsection{Paper Categorization Methdology}
\label{app:class}
In order to assess how geographic diversity changes at different stages of the NLP research and deployment, we use heuristics to sample papers according to the overarching phases associated with our Actor-Network. When developing heuristics, we prioritized precision over recall and manually sampled 30 papers from each group to confirm that all assignments were reasonable by the first authors judgement. For analysis, we restrict these to papers published the responsible NLP checklist, which requests authors report the hardware they used, was instituted in 2019 to avoid temporal shift between groups.

Our heuristics are as follows:
\begin{enumerate}
    \setlength{\itemsep}{0.025in}
    \item All papers from the Language Resources and Evaluation Conference (LREC) are used as a sample of Annotation and Benchmark creation. This results in 2447 papers which have a parseable PDF as our sample. While annotated resources are released via other venues, LREC is both a top venue and almost exclusively focuses on annotated resources and evaluation allowing us to keep with our principle of precision.
    \item All papers which use the phrase "our model", "our algorithm", or "our method" in the abstract of the work are used as a sample of Modeling and Methods papers. This results in 4940 papers which have a parseable PDF as our sample.
    \item Deployment is the phase which the ACL corpus is least representative. While many companies produce research papers, it is unclear which of these papers are ultimately deployed in products. We opt in favor of precision, counting only papers from workshops and or submission tracks with titles that include "Industry" or "In Practice". This results in 97 papers, which is a relatively small sample compared to all NLP deployment globally.
    \item Finally, to get a sense of hardware use in NLP, we collect all papers which mention "NVIDIA", "RTX", or "GeForce". This results in 2663 papers.    
\end{enumerate}

\section{Colonialism \& the Actor-Network}
In the Section \ref{sec:build}, we established our Actor-Network --- here we \textbf{trace the impacts of colonialism through our Actor-Network to material mediators in NLP} and ground them in our quantitative results.  
We start at the level of raw data, following ~\citet{decolonizingdata}, showing how digital infrastructure has been directly influenced by coloniality. Then, we discuss how this combines with annotation to multiply the power imbalances from raw data discrepancies. Finally, we discuss how increasing computing resources prevent stakeholders from re-appropriating NLP technology, further centralizing power. Throughout, we ground our analysis in results visualized in Figure \ref{ant_maps}.

\subsection{Colonialism \& Unlabeled Data Production}
Beyond coloniality of academic thought, colonial inequities are embedded in the language data we build on in NLP. \citet{whose} opens with a stark example of this --- the fiber optic cables connecting the globe are built on top of colonial trade routes\footnote{Also noted in \href{https://twitter.com/Abebab/status/1654523760408395777}{unpublished work by Abeba Birhane}.}~\citep{thorat2019colonial}. 
Digital enterprises were encouraged to prioritize quality of service for languages spoken by their users. As a result, languages with non-Latin scripts had poor early support, forcing speakers of many languages to romanize their language to be represented digitally~\citep{darwish-2014-arabizi, vanesch2019writing}. In a UNESCO report, \citet{prado2012language} noted that in 2012 Google supported "\emph{thirty European languages recognizes only one African language and no indigenous American or Pacific languages}". 

However, digital access and support of languages have expanded over time~\citep{van2014digital}. In the "Unlabeled Data Production" under Figure \ref{ant_maps}, we show that internet access increasingly is not defined by geographical discrepancies in access, with 80.5\% of users lying outside of Europe and British settler-colonial states. Even as access equalizes, the historical digital divide creates a "value lock" in the online digital archive.~\citep{bender2021dangers}. Recent corpora designed for multilingual language modeling skew significantly towards European languages~\citep{chung2023unimax}. This sets up internet users from the Global North as the most frequent voices describing all places and cultures~\citep{graham2014uneven, photoConcentration, elazar2023whats}.

Note that digital language data is not a priori positive resource~\citep{bird2020decolonising}. Even when many languages are represented in a dataset, they may reflect inherently colonial perspectives. \citet{ogunremi2023decolonizing} points to a clear example of this: religious texts, with deep ties to colonial evangelism, are often a major source of low-resource language data~\citep{christodouloupoulos2015massively, agic2019jw300}. Furthermore, many communities do not want digital archives or technology for their languages~\citep{bird2022local}, especially if it is controlled by an outside power. The first step of NLP work should be \texttt{to understand whether and how speakers see Language Technology applications as beneficial to their lives}.

\begin{figure*}[ht!]
\centering
   \begin{subfigure}[b]{0.325\textwidth}
   \includegraphics[width=1\textwidth]{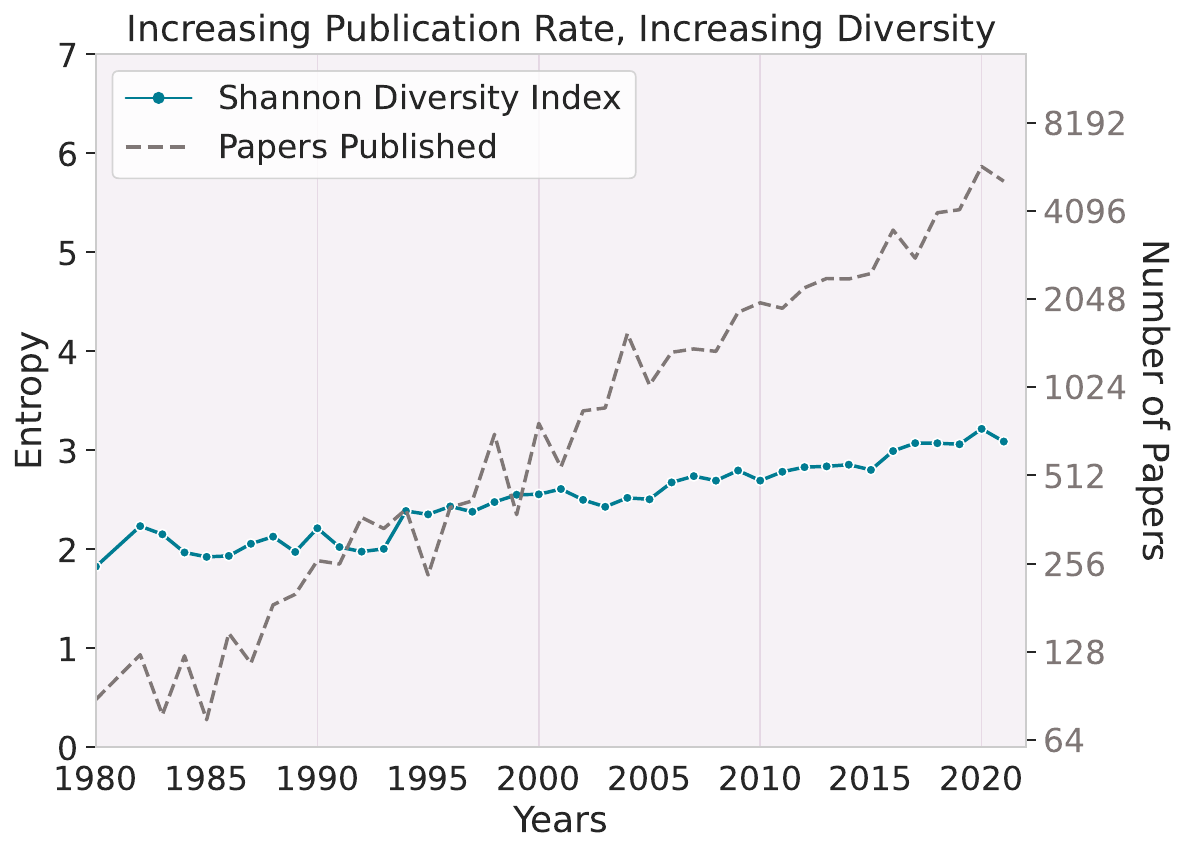}
   \caption{Linguistic diversity on the scale of the maximum and minimum possible entropy.}
   \label{fig:div} 
\end{subfigure}
\hfill
\begin{subfigure}[b]{0.65\textwidth}
       \includegraphics[width=1\textwidth]{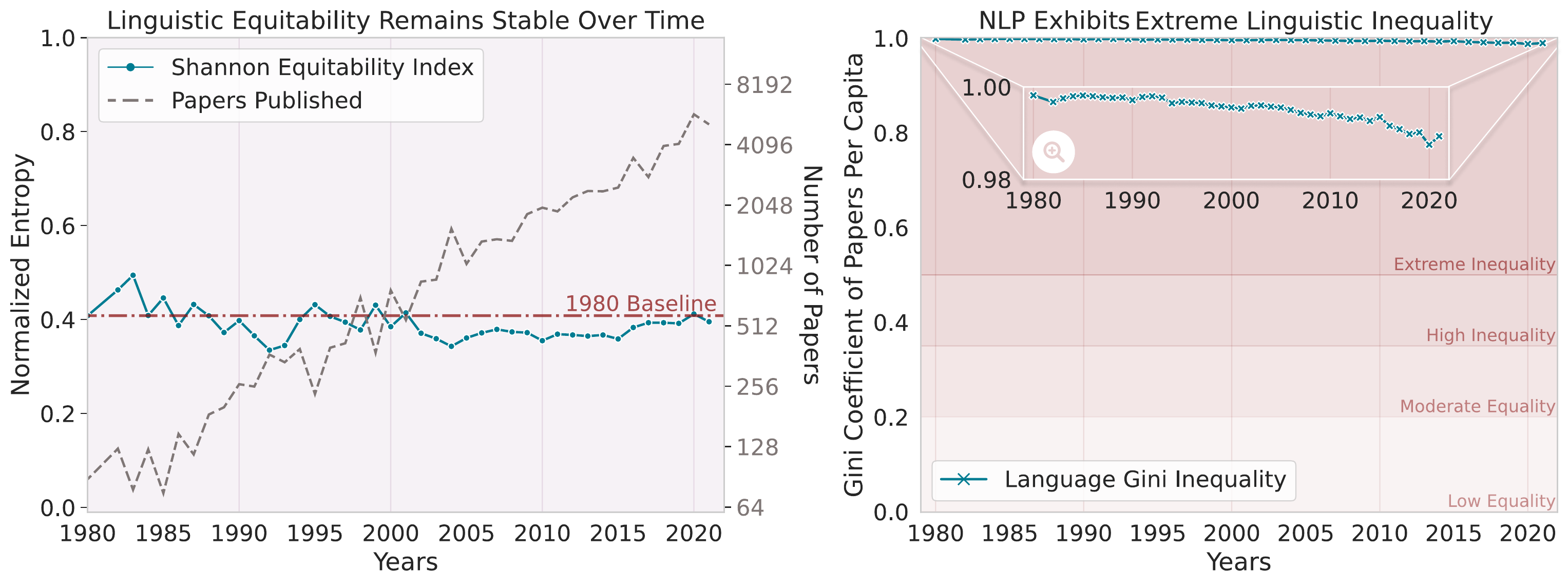}
   \caption{Linguistic equality, measured by both Shannon Equitability and Gini Inequality, on the scale of maximum and minimum equitability or inequality respectively.}
   \label{fig:equi}
\end{subfigure}

\caption{Diversity and equitability of languages mentioned in *CL publications for each year since the first Proceedings of the ACL. Metrics based on Shannon Entropy are compared to the number of papers published each year in NLP. See Section \ref{app:lang_mentions} for details on how we counted mentions in papers and computed each metric.}
\end{figure*}

\subsection{Annotation \& the Research Agenda}
As discussed in our development of the Actor-Network, the process of repeatedly resampling data magnifies its biases. In the "Annotation" section of Figure \ref{ant_maps}, we see that over 60.3\% of papers from LREC come from Europe alone. This likely influenced by the fact that LREC is organized by the European Language Resource Association LREC and has only once been hosted outside of Europe. The annotated resources produced for LREC  influence subsequent research agendas and further solidify the initial biases into the foundations of the field. 

For almost any interest area, English datasets are more likely to already exist to test many different modeling techniques or tasks~\citep{faisal-etal-2022-dataset, longpre2023data}. If a researcher's interests are truly "language-agnostic" or even if they have a slight preference to work on other languages, existing English resources may bias them to work on English to focus more directly on modeling~\citep{thakkar2022ml} and avoid annotation where the path to tangible results requires greater time and financial investment~\citep{sambasivan2021everyone}.

This process centers modeling in NLP on languages of the Global North where datasets already exist. Historically, NLP has termed systems "language-agnostic" even when the performance is only tested on English~\citep{bender2011achieving}. When systems are not tested on typologically and geographically diverse languages but are termed "language-agnostic" or simply "multilingual", it perpetuates the idea that all languages are substitutable. This is reductive, recreating colonial scientific tendencies to "\emph{[anatomize] and [melt] human entities as if they were so much inert matter}" \citep{said1978orientalism}. 

Positively, as shown in \citet{joshi-etal-2020-state} and reproduced for our corpus in Figure \ref{fig:div}, linguistic diversity\footnote{We follow \citet{joshi-etal-2020-state}, quantifying diversity using the Shannon Entropy of language mentions. Broadly, this usage is called the Shannon Diversity Index\citep{magurran2021measuring}.} in NLP is increasing. However, diversity metrics are negatively biased for small populations~\citep{konopinski2020shannon} and the increase coincides with a massive increase in the volume of NLP publishing. By comparison, Shannon Equitability, which normalizes the entropy used in \citet{joshi-etal-2020-state} by the maximum possible entropy~\citep{sheldon1969equitability}, has not significantly increased since the publication of the first ACL Proceedings. This inequality is clearly stated through the Gini Coefficient, a standard economic inequality metric, shown on the right in Figure \ref{fig:equi}: Inequality between the study of languages has never left the range of "extreme inequality\footnote{Based on Gini Inequality categories used for income inequality by the UN ILO~\citep{luebker2010inequality}}.

This challenges diversity alone as an indicator of equity, especially in regards to credit allocation. Data production for left-behind languages generally has a smaller audience of researchers than similar work done in English or other high-resource languages. This can be seen empirically through the relatively small clusters of publishing authors for most languages~\citep{joshi-etal-2020-state}. When researchers opt to develop tools for their own language rather than a high-resource language such as English, they are likely to receive less career reward for their work. While not purely driven by this, North American and European research receives systematically higher citations~\citep{citationgaps}. In the "Modeling" section of Figure \ref{ant_maps}, we see that while representation of China expands (29.4\% of Modeling papers, +25.9 from Annotation), South American (0.33\% of Modeling papers, -1.79 from Annotation) and African (0.28\% of Modeling papers, -0.26 from Annotation) representation dwindles to near-zero despite these regions making up 16.4\% of all internet users.

\begin{figure}[t]
\centering
\includegraphics[width=0.49\textwidth,trim={0cm 0.1cm 6.3cm 1cm},clip]{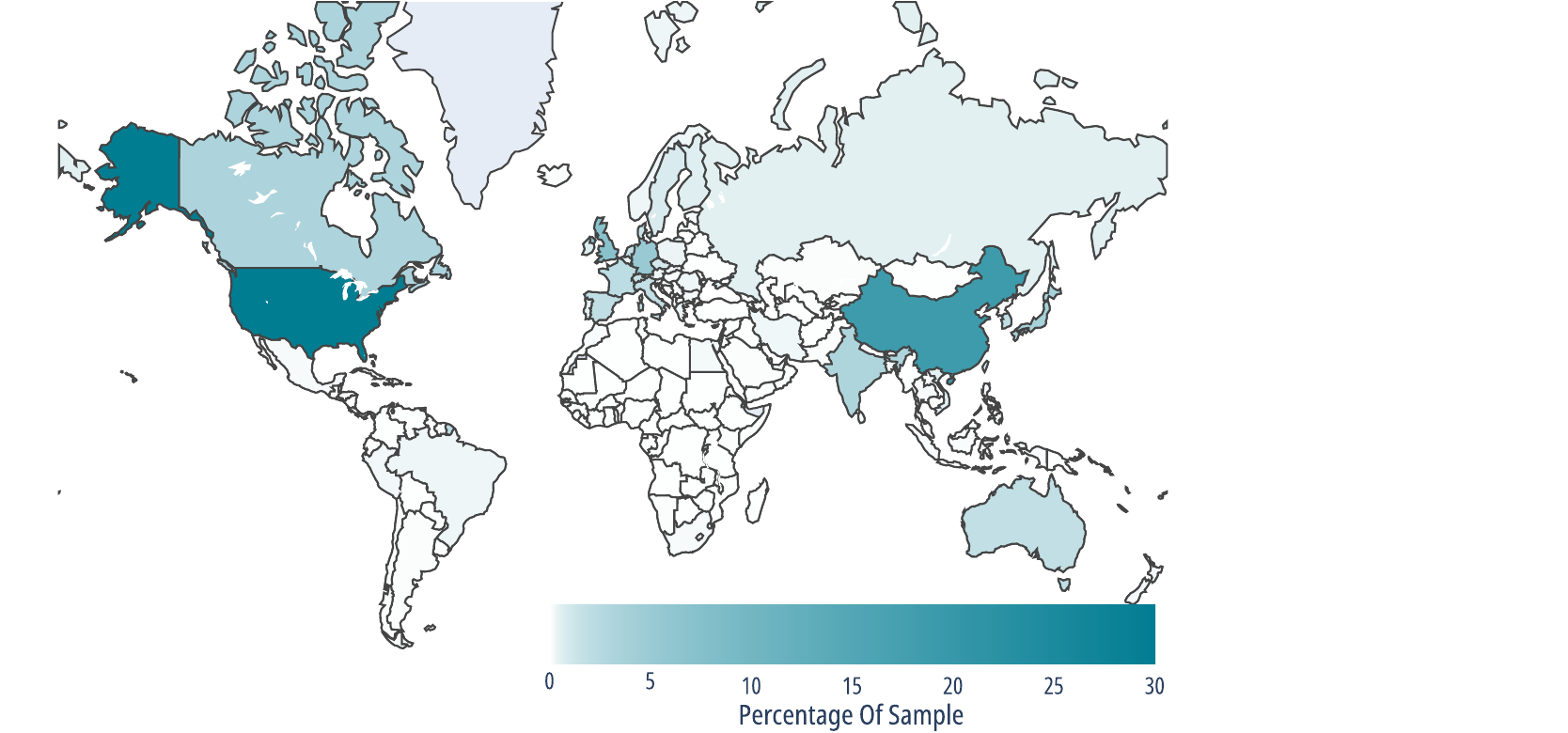}
\caption{Geographic density of papers which mention using NVIDIA GPUs. Only includes papers after the ACL added the Responsible NLP Checklist, which asks that all papers to report hardware used.}\label{fig:gpus}
\end{figure}

\subsection{Hardware \& the Control of Deployment}
Central to coloniality in NLP is the idea that stakeholders are often subject to NLP systems without the agency to change or reject them~\citep{bird2020decolonising, ogunremi2023decolonizing, decolonizingdata}. In the "Deployment" section of Figure \ref{ant_maps}, we approximate this using papers published in the Industry track and workshops focused on NLP "in practice", though admittedly NLP practitioners in the Global South may simply receive fewer rewards from published work in these venues. There are no papers from researchers or companies based in the Middle East, North Africa, Sub-Saharan Africa, or South America in these deployment focused venues. 

However, NLP is not the first colonial technology deployed across the globe and nations which suffered colonization have a history of adapting technologies used in service of colonialism into locally valuable tools, a process called appropriation~\citep{fanon1959dying, srinivasan2006indigenous, escobar2011sustainability}. This requires that people can own, modify, and reject technology according to their own value systems and desired usage. Recently, NLP has seen many appropriative initiatives that develop with and for historically colonized communities such as AmericasNLP~\citep{mager-etal-2018-challenges, mager-etal-2021-findings, ebrahimi-etal-2022-americasnli, ebrahimi-etal-2023-findings}, Masakhane~\citep{orife2020masakhane, nekoto-etal-2020-participatory, adelani-etal-2021-masakhaner, adelani2022few, adelani2022masakhaner}, and AI4Bharat~\citep{kakwani-etal-2020-indicnlpsuite, mhaske-etal-2023-naamapadam}. \citet{nekoto-etal-2020-participatory} is a gold standard in how this process can empower a community. Their work is led by community initiative from speakers of target low-resourced African languages and centers on providing agency to all actors rather than just the technologists involved. 

Appropriative efforts such as this can face a major material barrier --- the low-resource double bind~\citep{ahia-etal-2021-low-resource}. Even when data is accessible, state-of-the-art models often assume access to high-performance computing systems, such as the increasing reliance on Graphics Processing Units (GPUs). Access to even low-end GPUs is not universal, as we show in Figure \ref{fig:gpus}. Over 50\% of GPU usage in recent *CL publications is from just two countries, China and the United States. While both nations are large producers of NLP research in general, a paper which uses a GPU is 22.7\% more likely to be from these two nations than a randomly sampled paper in the same time period.

Some may argue that providing access through APIs reduces this barrier by providing access to models without fixed costs. However, this subjects the appropriative process to the limitations laid out by providers, who are based primarily in North America and Europe. The resulting systems are often poorly designed for the historically colonized, while undermining local innovation~\citep{birhane2020algorithmic}. With recent NLP models, we have already seen more costly pricing and weaker safety systems for low-resource language speaking communities~\citep{ahia2023languages, yong2023lowresource}.

\section{Case Studies In Coloniality}
\label{sec:case}
We present two case studies to illustrate that quantitative measures capture only a subset of the impacts of coloniality on technological artifacts. We first present an example of how the legacy of colonialism affects the Black population of the United States, showing that coloniality also applies to forms of colonizer languages that originate from colonized subjects. We then discuss how even within Multilingual NLP, common practices in NLP extend Western biases to new languages.

\subsection{Disparities for Minority English Dialects\\ \textit{\small{Key Material Actors: Digital Language, Datasets}}}

While the Black population of the United States are often not regarded as colonial subjects, Black people in the U.S. are subject to internal colonialism\footnote{\citet{gutiérrez_2004} defines internal colonialism as a system of exploitation within the oppressor state characterized by four key relationships of domination: the forced entry of the oppressed group to the dominant society; the transformation, or destruction of the value systems of the colonized; the administration of the colonized population by the oppressor nation; and racism towards the colonized population.}. During chattel slavery, Black people in the US were forced to adopt English and stop speaking indigenous African languages, mixing grammar patterns, pronunciations, and expression of various West African languages with 17th century British English \cite{winford2015origins}. This resulted in African American Vernacular English (AAVE) also called African American English (AAE) or African American Language (AAL). 

Enslaved Africans were barred from writing, creating a centuries-long oral tradition~\cite{winford2015origins}. In recent history, AAVE has been excluded from institutions such as media, literature, and education~\cite{jackson1997oakland}. This has directly created a lack of unlabeled textual resources for AAVE, except for recently collected social media data from Black users \cite{blodgett2017racial}. Even these datasets are limited as they are collected without consent or interaction with writers. As a result, AAVE is decontextualized creating risks of mockery and cultural appropriation from representations of AAVE available online \cite{ronkin1999mock, smokoski2016voicing}. However, they are often the only naturally occurring data to train and evaluate NLP models for AAVE~\cite{blodgett2017racial, jorgensen2016learning}.

When English datasets are annotated for AAVE, non-AAVE-speaking annotators may encode the lack of understanding of the language. Anti-Black stereotypes of aggressiveness, violence, and lack of intelligence, may also influence perceptions of AAVE and resulting annotations. A study from the Pew Research Center finds that only 6\% of Amazon Mechanical Turk workers are Black\footnote{https://www.pewresearch.org/internet/2016/07/11/turkers-in-this-canvassing-young-well-educated-and-frequent-users/}, making Black representation on the most widely used annotation platform disproportionately lower than the Black population of the U.S. (around 13\%). 

The systematic exclusion of AAVE from both unlabeled and annotated data has resulted in models that may not accurately represent or apply to the Black population, but are applied without contextualization. Multiple prior works have shown that AAVE is more likely to be falsely classified as hate speech or toxic speech \cite{harris2022exploring, sap2019risk, davidson2019racial, halevy2021mitigating} and as having negative sentiment in sentiment analysis \cite{groenwold2020investigating}. One downstream influence of this is the negative impact on the social media experiences of Black users, who are more likely to have their social media posts removed \cite{haimson2021disproportionate, camilleTikTok}. 

Limited access to representative unlabeled data and qualified human annotators makes improving or even identifying these issues challenging. While interventions have been attempted to mitigate biases towards AAVE in annotation~\citep{sap2019risk, ziems2022value, dacon2022towards} none are a cohesive solution, especially due to the regional and generational variation of AAVE \citep{hinton2000regional, king2020african}. Without fundamentally reshaping annotation procedures with feedback and data offered by AAVE speakers~\citep{mengesha2021don}, new systems will perepetuate the issue even if researchers following best practice mitigations.

Due to the power of the US as the dominant source of published NLP literature --- African American English has received greater visibility~\citep{jorgensen2016learning, blodgett2017racial, dorn-2019-dialect, groenwold-etal-2020-investigating, ziems2022value} within NLP compared to other similarly situated languages such as Jamaican Patois~\citep{armstrong-etal-2022-jampatoisnli}, Haitian Kreyol~\citep{ponti2020xcopa, lent-etal-2021-language, lent2022creole}, Australian Aboriginal English~\citep{zwarts2007statistical, ziems2023multivalue}. Furthermore, the Black communities that are most impacted are often not involved in AAVE NLP research. While some colonized groups have their languages excluded, others such as AAVE speakers, have language technologies imposed on them unilaterally. 

\subsection{English Centric Multilingualism \\ \textit{\small{Key Material Actors: Datasets, Algorithms, Hardware}}}

Multilingual Language Models ~\citep{mbert, xlmr, mt5-xue} have successfully driven an increased research interest in NLP beyond English~\cite{joshi-etal-2020-state}. The growing popularity of such models has itself motivated a wave of cross-lingual benchmarks which annotate a single task over several languages~\citep{conneau2018xnli, yang-etal-2019-paws, nivre2018universal, Pan2017, artetxe2020cross, Lewis2020mlqa, Clark2020tydiqa, zweigenbaum2018overview, Artetxe2019massively, li-etal-2021-mtop, fitzgerald2022massive}. 

When researchers look to train multilingual systems, they usually adopt algorithms developed for the data-rich English context. These systems have data requirements far beyond what any single human consumes~\citep{warstadt2023papers}. Furthermore, reaching new performance heights seems to require exponentially more data with each new model~\citep{kaplan2020scaling, hoffmann2022training}. In pursuit of data to meet this scale, multilingual models often sample as much unlabeled data as possible from the internet~\citep{kreutzer-etal-2022-quality}. The impact of coloniality on unlabeled data production means that Western European languages, especially English, are over-represented compared to their relative population size~\citep{chung2023unimax}. By this process, even multilingual models are demonstrably skewed towards Western, and especially American, values and phrasing ~\citep{johnson2022ghost, arora-etal-2023-probing, naous2023having, papadimitriou-etal-2023-multilingual, durmus2023measuring}.

Ideally, benchmark evaluations would capture cultural discrepancies. However, data curators often translate existing annotated data~\citep{artetxe-etal-2020-translation}, likely aiming to extend an English-centric resource with accepted value to new languages. These datasets may cover many languages, but the names, places, events, and other cultural concepts are transferred into the evaluation for new languages. As an example, out of the 47 Wikipedia articles in the XQuaD~\citep{artetxe-etal-2020-translation} test set: 13 are primarily about North America (e.g. Super Bowl 50), 11 on Europe (e.g. Scottish Parliament), 2 on Asia (Yuan Dynasty \& Genghis Khan), 1 on Australia (the state of Victoria), 1 on Africa (Kenya), 1 on Latin America (Amazon Rainforest), 1 on the Middle East (Islamism), and the remaining 17 are location agnostic (e.g. Oxygen). 

This is common practice: 8 out of 12 tasks across in the widely used multi-task XTREME~\citep{xtreme} and XTREME-R~\citep{xtremer} benchmarks use data translated from English. However, such datasets may struggle to capture cultural misalignment and may in fact reward systems which replicate, especially in the range of tasks where cultural understanding has been shown to influence accuracy~\citep{mohammad2016translation, smith-etal-2016-well, asai-etal-2021-xor, bauer-etal-2023-social, lee-etal-2023-hate, akinade-etal-2023-varepsilon}. 

In this case, the data requirements of existing algorithms influence researchers to sample increasing amounts of Western data and the cultural biases of datasets underestimate the negative effects of this process. While real speakers of these languages may recognize this cultural misalignment, hardware constraints may inhibit them from correcting it in existing models. In order to change this, we must address not only social issues, but also technical challenges in data requirements and hardware requirements.

\section{Looking Forward}
We have shown that the power dynamics of NLP become increasingly stark across colonial boundaries as the broader social inequities of coloniality are exacerbated by NLP specific technological barriers. These further prevent the historically colonized from developing empowering interventions. This inequality of choice~\citep{sen1995inequality, sen2001development} leaves the historically colonized subject to the choices of technologists in the Global North. In our analysis, two of these barriers stand out as technical challenges where researchers can expand agency in the NLP actor network listed below.

\textbf{Reducing the low-resource double bind}~\citep{ahia-etal-2021-low-resource} is a key short term focus. Hardware limitations create major barriers for those in the Global South to appropriate language technology for their own use~\citep{aji-etal-2022-one}. Appropriation of technology~\citep{dix2007designing} is a partial solution that is well-established in cultures that survived colonialism, such as Jugaad in India~\citep{rangaswamy2013understanding} or Gambiarra in Brazil~\citep{boufleur2006questao}. Researchers who view themselves as "purely technical" can enable appropriation of existing technology according to local value systems without forcing technological solutionism by developing more efficient methods. The expansion of agency and choice can itself be seen as technological development~\citep{sen2001development, kleine2009ict4what}, even if the outcome is the rejection of language technologies by affected stakeholders. Notably, appropriation alone is insufficient as it is an act of survival under oppressive power structures rather than an indicator of equality~\citep{bar2016mobile}.

\textbf{Methods that reduce the reliance on unfathomably large datasets} to produce high-quality language technology, such as \citet{adelani2022few} and \citet{ogunremi-etal-2023-mini}, help break the inheritance of their problematic foundations~\citep{birhane2021large}. Data efficient learning allows curators to carefully design datasets in accordance with community-aligned values, rather than simply minimizing the biases online data. Creating room for value-centered curation is key to re-imagining datasets as a rudder which helps direct NLP as a field~\citep{bommasani2022evaluation} rather than as a measure of opaque notions of progress~\citep{raji2021ai}.

\textbf{Technical work is not enough}; we echo the calls of prior works~\citep{ogunremi2023decolonizing, bird2020decolonising, schwartz-2022-primum, bird2022local} that the NLP research community to make room for methodologies outside colonial traditions~\citep{smith1999decolonizing} and engage in participatory research~\citep{nekoto-etal-2020-participatory}. This may sometimes require more time for discussion to resolve conflict between viewpoints. However, this process is essential both ethically and to build systems which are effective in nuanced global contexts~\citep{nkemelu2022tackling}.

\textbf{Achieving linguistic diversity in NLP research is not, in itself, inclusion}. Stakeholders from the Global South are often involved as annotators for their languages, but excluded from the financial or reputational gains that authors receive from high-profile NLP projects~\citep{shmueli-etal-2021-beyond}\footnote{Reports on exploitation of annotators in the Global South for NLP in the \href{https://www.technologyreview.com/2022/04/20/1050392/ai-industry-appen-scale-data-labels/}{MIT Tech Review}, \href{https://time.com/6247678/openai-chatgpt-kenya-workers/}{TIME}}, and the \href{https://www.wsj.com/articles/chatgpt-openai-content-abusive-sexually-explicit-harassment-kenya-workers-on-human-workers-cf191483}{Wall Street Journal}. This perpetuates colonial methodologies in research: "research projects are designed and carried out with little recognition accorded to the people who participated"~\citep{smith1999decolonizing}. Participatory research is not a panacea for the malcontents of NLP if it does not provide agency and ownership of the language technology developed to the stakeholders subject to resulting systems~\citep{people}. Beyond technosolutionism, we must consider and discuss with stakeholders when NLP may exacerbate harms. 

Our code of ethics binds us to, regardless of intent, "ensure that all harm is minimized"~\citep{gotterbarn2018acm}\footnote{The ACL officially follows the \href{https://www.aclweb.org/portal/content/acl-code-ethics}{ACM Code of Ethics}.}. By understanding the ties between NLP and colonial ideals, we see that doing so is an active process of undoing the ways in which harm is inextricably tied to the foundations of our field.

\section{Limitations}

In developing our actor network, we focus on the relationships which exist pervasively across NLP. As the NLP research community is large with many focus areas, this generalized framework is likely to omit important actors which are relevant within specific contexts of NLP. When applying ANT to specific NLP tasks, we recommend using our framework for common elements, but moving towards defining concrete context specific key actors.

Furthermore, due to the expansiveness of European colonialism, coloniality has affected many diverse groups. Our work looks to prior work across these perspectives, but readers should not take our work as an indication that the desires of these communities are homogeneous. As we explore in our case study of African American Vernacular English, for some communities, language technologies are neither desired nor beneficial. Inclusion in the actor network should also be reflected as the right to reject language technology entirely. 

Finally, while our work focuses on the material expressions of coloniality, this is in addition to, not suggesting the lack of, ideological coloniality highlighted in prior work. This work was inspired by prior works by \citet{ogunremi2023decolonizing}, \citet{mohamed2020decolonial}, \citet{schwartz-2022-primum}, \citet{bird2020decolonising} \citet{mager2023ethical}, and \citet{decolonizingdata}. This focus should not be taken to indicate that ideological coloniality has been solved in NLP. Ideological coloniality continues to be pervasive and serves to increase material inequities. 

\section{Ethical Considerations}
This work uses the term "the historically colonized" to draw attention to the fact colonialism continues to affect those who are not presently under an official colonial regime. The use of this term may give the false impression that we live in a postcolonial context. The peoples of Anguilla, Bermuda, the Virgin Islands, the Cayman Islands, Malvinas, Montserrat, Saint Helena, Turks and Caicos, Gibraltar, American Samoa, French Polynesia, Guam, New Caledonia, Pitcairn, and Tokelau are still colonized. Drawing boundaries between marginalized communities and the dominant perspectives of NLP has the negative impact of othering already marginalized communities. 

In clustering languages according to their regional origin, Figure \ref{fig:headline} omits the fact that languages are adapted and transformed beyond their origins. Furthermore, we rely on Ethnologue for this information, a resource managed by SIL International, for this information. SIL itself is an organization with deep colonial roots~\citep{delvalls1978instituto}. While these combine to paint an incomplete picture, we do so to highlight that the major presence of researchers in nations with indigenous communities, such as the US, Canada, Australia, and Aotearoa (New Zealand), does not lead to representation of the indigenous languages of those nations.

Notably, we diverge from some prior works in one notable aspect, our lack of the use of the word "decolonizing" or "decolonial". We do so not because we believe that coloniality should not be dismantled in our field, but instead because these terms have a tendency to soften tangible requests for land return~\citep{tuck2012decolonization}.

\section{Acknowledgements}
We are grateful to Azure Zhou, Caleb Ziems, Dan Jurafsky, Dora Zhao, Irene Solaiman, Michael Li, Myra Cheng, Omar Shaikh, Pooja Casula, Pratyusha Ria Kalluri, Sachin Pendse,  Tol\'{u}l\d{o}p\d{\'{e}} \`{O}g\'{u}nr\d{\`{e}}m\'{i}, Tony Wang, Yanzhe Zhang, and Zhehao Zhang for feedback and suggestions at different stages of this work.
\bibliography{emnlp2023}
\bibliographystyle{acl_natbib}

\appendix

\end{document}